# Temporal-Spatial Feature Extraction Based on Convolutional Neural Networks for Travel Time Prediction

Chi-Hua Chen

College of Computer and Data Science, Fuzhou University, China

*Corresponding author: Chi-Hua Chen (e-mail: chihua0826@gmail.com).

**Abstract**

In recent years, some traffic information prediction methods have been proposed to provide the precise information of travel time, vehicle speed, and traffic flow for highways. However, big errors may be obtained by these methods for urban roads or the alternative roads of highways. Therefore, this study proposes a travel time prediction method based on convolutional neural networks to extract important factors for the improvement of traffic information prediction. In practical experimental environments, the travel time records of No. 5 Highway and the alternative roads of its were collected and used to evaluate the proposed method. The results showed that the mean absolute percentage error of the proposed method was about 5.69%. Therefore, the proposed method based on deep learning techniques can improve the accuracy of travel time prediction.

## 1. Introduction

In recent years, the growing travel demands have led to the increasing number of vehicles. For instance, the number of vehicles is increased from 6,876,515 to 7,842,423 between 2010 and 2016 [1]. Therefore, the traffic congestion may occur frequently, especially in peak hours (e.g., commute hours and holidays). The numbers of vehicles driven in No. 5 Highway in 2010 and 2016 were collected and showed in Table 1 [2]. This study used t-test to evaluate the practical data for analysing the number of vehicles. The results indicated that there was a significant difference (i.e., the p-value is between 0.00048 and 0.00518) between the numbers of vehicles on weekends and ordinary days. Furthermore, the number of vehicles increases 1.14 times in Taiwan from 2010 to 2016. Therefore, this study assumed that the number of vehicles in 2010 as the parameter c, and the estimated number of vehicles in 2016 can be calculated as $c \times 1.14$. The results indicated that there was a significant difference (i.e., the p-value is between 0.00052 and 0.00109) between the estimated number and practical number of vehicles in 2016, so the travel demands from Taipei County to Yilan County grew speedily. The detection and prediction of traffic congestion in No.

5 Highway in Taiwan are an important issue.

Therefore, governments built several intelligent transportation systems (ITS)(e.g., advanced traffic management system and advanced traveller information system) to monitor and provide real-time traffic information for user decision-making. Although the real-time traffic information (e.g., vehicle speed, traffic flow, travel time, etc.) can be detected by vehicle detectors [3] and eTag (electronic tag) detectors [4], the future traffic information in dynamical environments cannot be obtained. Therefore, some study proposed linear regression [5, 6], logistic regression [7], and neural networks (NN) [8-10] to analyse real-time and historical traffic information for predicting future traffic information. Although these methods can solve the linear and non-linear problems, overfitting problems may be occurred [11-13]. Therefore, the errors of traffic information prediction were larger than 20% for urban roads and the alternative roads of highways [14]. Therefore, this study proposes a travel time prediction method based on deep learning techniques [15] and convolutional neural networks (CNNs) [16-19] to extract features based on several filters for the improvement of traffic information prediction.

Table 1. The numbers of vehicles driven in No. 5 Highway in 2010 and 2016

| Road segments | Direction | The number of vehicles in 2010 (Practical values) (i.e., the parameter $c$) | | | The number of vehicles in 2016 (Estimated values) (i.e., $c \times 1.14$) | | | The number of vehicles in 2016 (Practical values) | | |
|---|---|---|---|---|---|---|---|---|---|---|
| | | Saturday | Sunday | Ordinary days | Saturday | Sunday | Ordinary days | Saturday | Sunday | Ordinary days |
| From Shiding To Pinglin | Southward | 33,370 | 28,183 | 18,864 | 38,042 | 32,129 | 21,505 | 40,364 | 33,476 | 26,306 |
| From Pinglin To Toucheng | Southward | 32,047 | 26,633 | 18,021 | 36,534 | 30,362 | 20,544 | 38,278 | 30,405 | 24,799 |
| From Toucheng To Yilan | Southward | 26,638 | 20,988 | 17,222 | 30,367 | 23,926 | 19,633 | 31,818 | 23,863 | 21,944 |
| From Yilan To Luodong | Southward | 20,609 | 17,109 | 15,019 | 23,494 | 19,504 | 17,122 | 24,637 | 19,410 | 17,878 |
| From Luodong To Yilan | Northward | 17,140 | 17,667 | 14,034 | 19,540 | 20,140 | 15,999 | 20,050 | 19,179 | 17,286 |
| From Yilan To Toucheng | Northward | 22,518 | 25,179 | 16,912 | 25,671 | 28,704 | 19,280 | 26,938 | 27,899 | 21,841 |
| From Toucheng To Pinglin | Northward | 26,264 | 32,215 | 17,840 | 29,941 | 36,725 | 20,338 | 32,089 | 34,137 | 25,053 |
| From Pinglin To Shiding | Northward | 27,642 | 34,513 | 18,768 | 31,512 | 39,345 | 21,396 | 34,606 | 39,931 | 26,538 |

The remaining of the paper is organized as follows. The literature reviews of traffic information prediction methods are discussed in Section 2. The proposed traffic information prediction system and method based on CNNs are clearly illustrated in Section 3. In Section 4, experimental environments are presented, and the practical results are given and discussed for the evaluation of the proposed method. The conclusion and future work are provided in Section 5.

## 2. Related Work

This study describes and defines traffic characteristics in Subsection 2.1. Traffic information collection and prediction methods are discussed and compared in Subsections 2.2 and 2.3.

### 2.1. Traffic Characteristics

Traffic characteristics include traffic flow, vehicle speed, vehicle density, and travel time. Vehicle speed can be defined as time mean speed and space mean speed by different measurement methods. Each traffic characteristic is illustrated and defined in the following subsections.

#### 2.1.1. Traffic Flow

Traffic flow is defined as $Q$ car/hr to show the number of vehicles driven through a specific location. For instance, traffic information is periodically reported in each five-minute in Taiwan. In Figure 1, when there were 97 vehicles driven through Location 1 from 6:00 a.m. to 6:05 a.m. on July 11 in 2017, the traffic flow of Location 1 was reported as 1,164 car/hr (i.e., 97 car/5min).

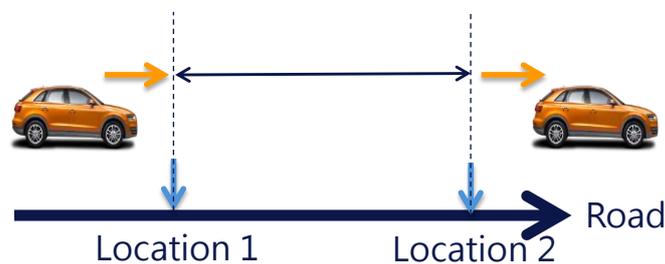

Figure 1. Traffic characteristics

#### 2.1.2. Time Mean Speed

Time mean speed (TMS) is defined as $U_t$ km/hr to show the average of instantaneous speeds of vehicle driven through a specific location. For instance, when the mean instantaneous speed of 97 vehicles driven through Location 1 from 6:00 a.m. to 6:05 a.m. on July 11 in 2017 was 88 km/hr, the TMS of Location 1 was reported as 88 km/hr at 6:05 a.m. (shown in Figure 1).

#### 2.1.3. Space Mean Speed

Space mean speed (SMS) is defined as $U_s$ km/hr to show the average of movement speeds of vehicle driven from one location (i.e., origin) to another location (i.e., destination). For instance, there are 73 vehicles driven from Location 1 to Location 2 between 6:00 a.m. and 6:05 a.m. on July 11 in 2017 (shown in Figure 1). When the mean movement speed of these vehicles driven from Location 1 to Location 2 was 87 km/hr, the SMS between Location 1 and Location 2 was reported as 87 km/hr at 6:05 a.m.

### 2.1.4. Vehicle Density

Vehicle density is defined as $K$ car/km to show the number of vehicles driven from one location (i.e., origin) to another location (i.e., destination). However, it is difficult to get the precise vehicle density, especially in urban roads. Therefore, vehicle density is usually estimated by a vehicle detector in accordance with traffic flow and TMS (shown in Equation (1)). For instance, the traffic flow and TMS were reported as 1,164 car/hr and 88 km/hr, respectively. Therefore, vehicle density was estimated as 13 car/km (shown in Equation (2)) and reported at 6:05 a.m. on July 11 in 2017.

$$K = \frac{Q}{U_t} \tag{1}$$

$$K = \frac{Q}{U_t} = \frac{1164}{88} \fallingdotseq 13 \tag{2}$$

### 2.1.5. Travel Time

Travel time is defined as $T$ hr to show the average of time differences of vehicles driven from one location (i.e., origin) to another location (i.e., destination). In Figure 1, the geographic distance from Location 1 to Location 2 is $D$ km, and the value of $D$ is 3 in this case. Therefore, the travel time can be estimated by Equation (3) in accordance with the geographic distance and SMS from Location 1 to Location 2. In this case, travel time from Location 1 to Location 2 was estimated as 0.0345 hr (shown in Equation (4)) and reported at 6:05 a.m. on July 11 in 2017.

$$T = \frac{D}{U_s} \tag{3}$$

$$T = \frac{D}{U_s} = \frac{3}{87} \fallingdotseq 0.0345 \tag{4}$$

### 2.2. Traffic Information Collection Method

This subsection presents main traffic information collection methods including vehicle detectors and eTag detectors.

### 2.2.1. Vehicle Detectors

The vehicle detectors can detect the traffic flow and TMS based on the techniques of sensing coils or cameras. These sensing coils or cameras are built at the intersection of roads, and an event can be triggered for traffic information detection when a vehicle is driven through the intersection of roads. Furthermore, the vehicle density can be estimated according to the detected traffic flow and TMS.

This study collected the traffic information from a vehicle detector in No. 1 Highway in Taiwan during February of 2008 to discuss the relationships among traffic flow, vehicle speed, and vehicle density. In this case, the traffic information (e.g., traffic flow, vehicle speed, and vehicle density) was periodically reported in each five-minute, and each record was presented as a blue dot in Figures 2, 3, and 4. For instance, the blue dot in the red circle in Figure 2 was reported to show that the traffic flow and TMS were 2028 car/hr and 12 km/hr at 19:40 p.m. on 27 February 2008, respectively.

Figure 2 showed the effects of traffic flow on vehicle speed. When the traffic condition was free flow, TMS was about 90 km/hr and traffic flow was lower. However, when the traffic condition was congestion in the ahead of Location 1 in Figure 1, the traffic flow and TMS of Location 1 were lower and slower (shown in the red circle in Figure 2), respectively.

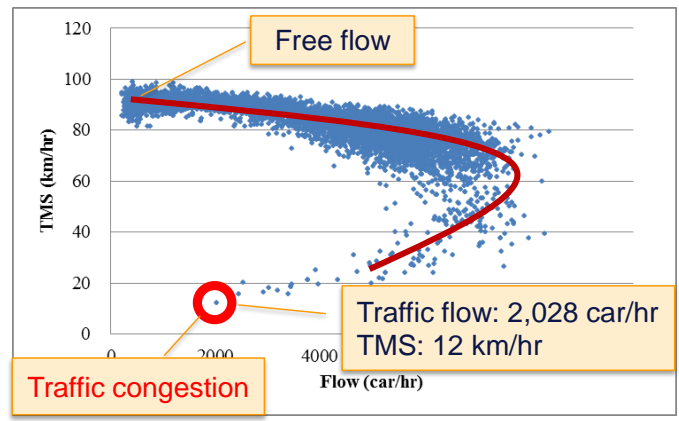

Figure 2. The effects of traffic flow on vehicle speed

Figure 3 showed the effects of vehicle density on vehicle speed. In the case of free flow, vehicle density was lower, and TMS was higher. Furthermore, in the case of traffic congestion, the vehicle density was higher, and TMS was lower. The relation of vehicle density and vehicle speed could be shaped as a backslash.

Figure 4 showed the effects of vehicle density on traffic flow. The relation of vehicle density and vehicle speed could be drawn as an inverse of U-shape. When the traffic condition was free flow, the both of vehicle density and traffic flow were lower.

In the case of traffic condition was congestion in the ahead of Location 1 in Figure 1, the vehicle density was higher, but traffic flow was lower (shown in the red circle in Figure 4).

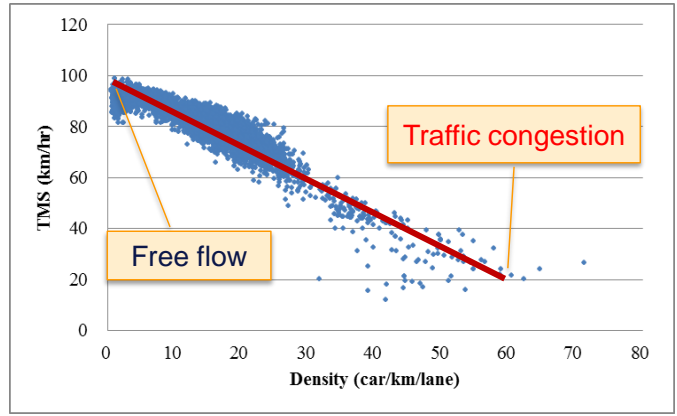

Figure 3. The effects of vehicle density on vehicle speed

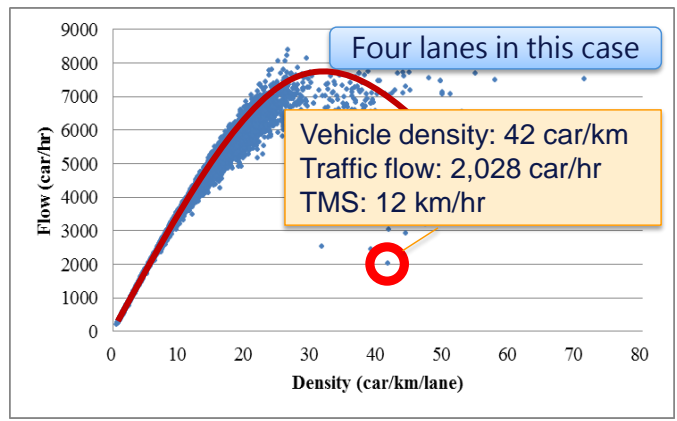

Figure 4. The effects of vehicle density on traffic flow

### 2.2.2. eTag Detectors

The eTag detectors based on techniques of electronic toll collection were built in the whole highways in Taiwan for traffic information collection. The 94% of vehicles were equipped with an eTag in Taiwan. The eTag detectors can detect the movement of each vehicle equipped an eTag from one location (i.e., origin) to another location (i.e., destination) and generate the travel time of this trip. Therefore, eTag detectors can obtain the travel time of each road segment, and the SMS of each road segment can be estimated by Equation (5). For instance, eTag detectors were built at Locations 1 and 2 in Figure 1, respectively. When a vehicle moved through Location 1, the eTag detector at Location 1 can detect the ID of the vehicle (i.e., $ID_1$) at Time $t_1$. Furthermore, the vehicle keeps moving through Location 2, and the eTag detector at Location 2 can detect $ID_1$ at Time $t_2$. Therefore, the travel time of the vehicle from Location 1 to Location 2 can be measured as $(t_2 - t_1)$ (i.e., travel time $T$). Moreover, the geographic distance between Location 1 and Location 2 was predefined as *D* km. Therefore, Equation (5) can be used to obtain the information of SMS from Location 1 to Location 2 according to the geographic distance and travel time.

$$U_s = \frac{D}{T} \tag{5}$$

### 2.2.3. Summary

In accordance with the relationships among traffic flow, vehicle speed, and vehicle density, the traffic information of a specific location may be affected by the traffic condition in the ahead of the location. Therefore, adjacent road segments should be considered for traffic information prediction. Furthermore, although vehicle detectors can obtain the information of traffic flow and TMS, travel time and SMS which cannot be obtain by vehicle detectors are important factors for users and governments. Therefore, eTag detectors are applied to detect real-time travel time and SMS, and the traffic information of adjacent road segments are considered in the proposed method for the improvement of travel time prediction.

## 2.3. Traffic Information Prediction Method

The previous traffic information prediction methods which include (1) unweighted average method, (2) linear regression and logistic regression, (3) neural network, and (4) convolutional neural networks are presented in the following subsections.

### 2.3.1. Unweighted Average Method

The unweighted average method analyses the mean of historical SMS in a specific road segment for obtain the predicted SMS. For instance, there are *n* SMS records of Road Segment 2 (shown in Figure 5) in historical database. The SMS time of Road Segment 2 at Time *i* is defined as $U_{s,2,i}$, and the SMS of Road Segment 2 at Time (*n*+1) can be predicted as $U_{s,2,n+1}'$ by Equation (6).

$$U_{s,2,n+1}' = \frac{\sum_{i=1}^{n} U_{s,2,i}}{n} \tag{6}$$

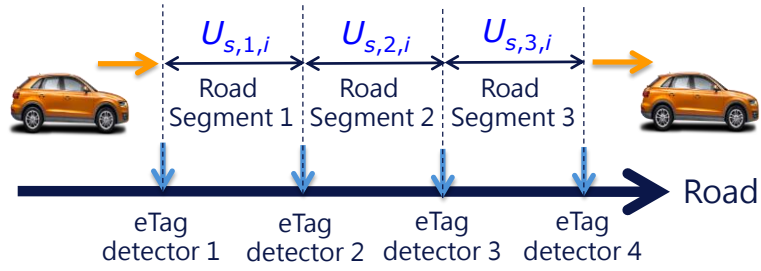

Figure 5. The space mean speeds of adjacent road segments

Although the unweighted average method is simple and fast to predict the SMS of target road segment (i.e., Road Segment 2 in Figure 5), the traffic conditions of adjacent road segments (e.g., Road Segments 1 and 3 in Figure 5) are not considered in this method. However, the traffic information of target road segment may be affected by traffic conditions of adjacent road segments. Therefore, the unweighted average method is unsuitable for dynamical environments.

### 2.3.2. Linear Regression and Logistic Regression

Some studies proposed linear regression and logistic regression methods to analyse the traffic conditions of adjacent road segments. The SMSs of adjacent road segments were used as input parameters and adopted into the traffic information prediction models based on regression methods [5-7]. For instance, the SMS of Road Segment 2 in Figure 5 could be predicted by the regression model in Figure 6. In this model, the input parameters included the SMS of target and adjacent road segments (i.e., $U_{s,1,i}$, $U_{s,2,i}$, and $U_{s,3,i}$) at Time $i$, and the output parameter included the SMS of target road segment (i.e., $U_{s,2,i+1}$) at Time ($i$+1). Historical SMS records could be used to train the weights (i.e., $w_{s,1}$, $w_{s,2}$, and $w_{s,3}$) and bias (i.e., $b_{s,2}$), and Equations (7) and (8) could be used to obtain the prediction results of linear regression and logistic regression, respectively.

$$U_{s,2,i+1} = \left( \sum_{j=1}^{3} U_{s,j,i} \times w_{s,j} \right) + b_{s,2} \tag{7}$$

$$U_{s,2,i+1} = S\!\left( \left( \sum_{j=1}^{3} U_{s,j,i} \times w_{s,j} \right) + b_{s,2} \right),$$
$$\text{where } S(x) = \frac{1}{1+e^{-x}} \tag{8}$$

Although the traffic conditions of adjacent road segments are considered by linear regression and logistic regression methods, the inputs of regression model are assumed as independent parameters. Therefore, big errors of traffic information prediction may be generated when the dependencies of input parameters are significant. For example, the traffic information of Road Segment 1 may be affected by the traffic condition of Road Segment 3. Therefore, the regression methods are not the optimal solutions.

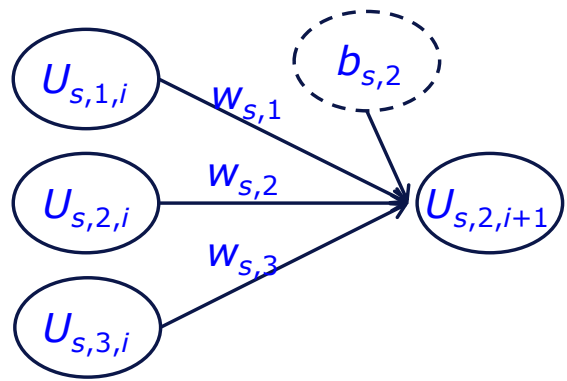

Figure 6. The model of regression

### 2.3.3. Neural Network

For analysing the dependencies and interactions of input parameters, some studies proposed neural networks including an input layer, hidden layers, and an output layers (shown in Figure 7). The hidden layers could be used to analyse the dependencies and interactions of input parameters for improvement of traffic information prediction [8-10]. For instance, the SMS of target and adjacent road segments (i.e., $U_{s,1,i}$, $U_{s,2,i}$, and $U_{s,3,i}$) at Time $i$ were adopted as the neurons in the input layer in Figure 7, and the SMS of target road segment (i.e., $U_{s,2,i+1}$) at Time ($i+1$) was adopted as the neuron in the output layer for predicting the SMS of Road Segment 2. In this model, two neurons were built in the hidden layer, and the values of them could be calculated by Equations (9) and (10), respectively. Furthermore, the outputs could be estimated by Equation (11) in accordance with the neurons in the hidden layer. The gradient descent method and historical records were used to train each weight and bias in this model for traffic information prediction.

$$U_{s,2,i,h1} = S\left(\left(\sum_{j=1}^{3} U_{s,j,i} \times w_{s,j,h1}\right) + b_{s,h1}\right), \quad (9)$$

$$\text{where } S(x) = \frac{1}{1+e^{-x}}$$

$$U_{s,2,i,h2} = S\left(\left(\sum_{j=1}^{3} U_{s,j,i} \times w_{s,j,h2}\right) + b_{s,h2}\right), \quad (10)$$

$$\text{where } S(x) = \frac{1}{1+e^{-x}}$$

$$U_{s,2,i+1} = S\left(\left(\sum_{k=1}^{2} U_{s,2,i,h_k} \times w_{s,h_k}\right) + b_{s,2}\right), \quad (11)$$

$$\text{where } S(x) = \frac{1}{1+e^{-x}}$$

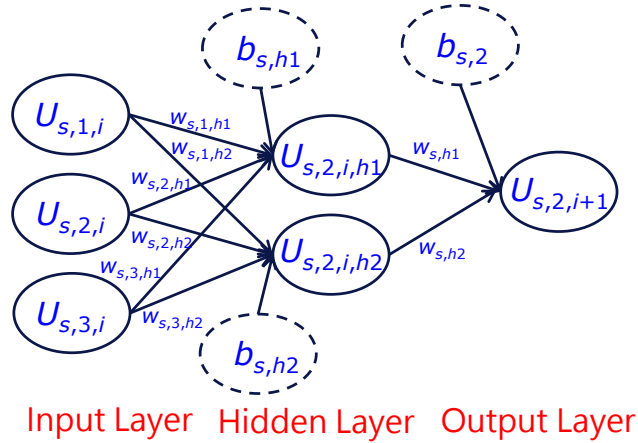

Figure 7. The model of neural network

Although neural networks can analyse the dependencies and interactions of traffic conditions of adjacent road segments, the prevention of overfitting problems is not considered [11-13, 20]. In training stage, the weights in a fully-connected neural networks are trained for historical cases. Therefore, these neural networks cannot provide a generalized solution for the interaction effects of space factors, and big errors of traffic information prediction may be generated in testing stage and execution stage.

### 2.3.4. Convolutional Neural Network

Convolutional neural networks included convolutional layers, pooling layers, and a neural network [16-19] (shown in Figure 8). Each component of CNNs is presented in the following paragraphs.

**(1). Convolutional layers**

In convolutional layers, several filters (i.e., the sets of weights) can be generated by the gradient descent method to extract features and build several feature maps with partially connections among neurons. The fully connections between two layers are considered in a normal neural network mentioned in Subsection 2.3.3. However, a neural network with fully connections is assumed that the dependencies of each two-neuron are significant. Therefore, big errors may be obtained when the neural network with the independencies of some input parameters. Therefore, convolutional layers with partially connections can only perform to analyse the traffic conditions of adjacent road segments, and the traffic information of distant road segments is not considered in convolutional layers for the improvement of traffic information prediction.

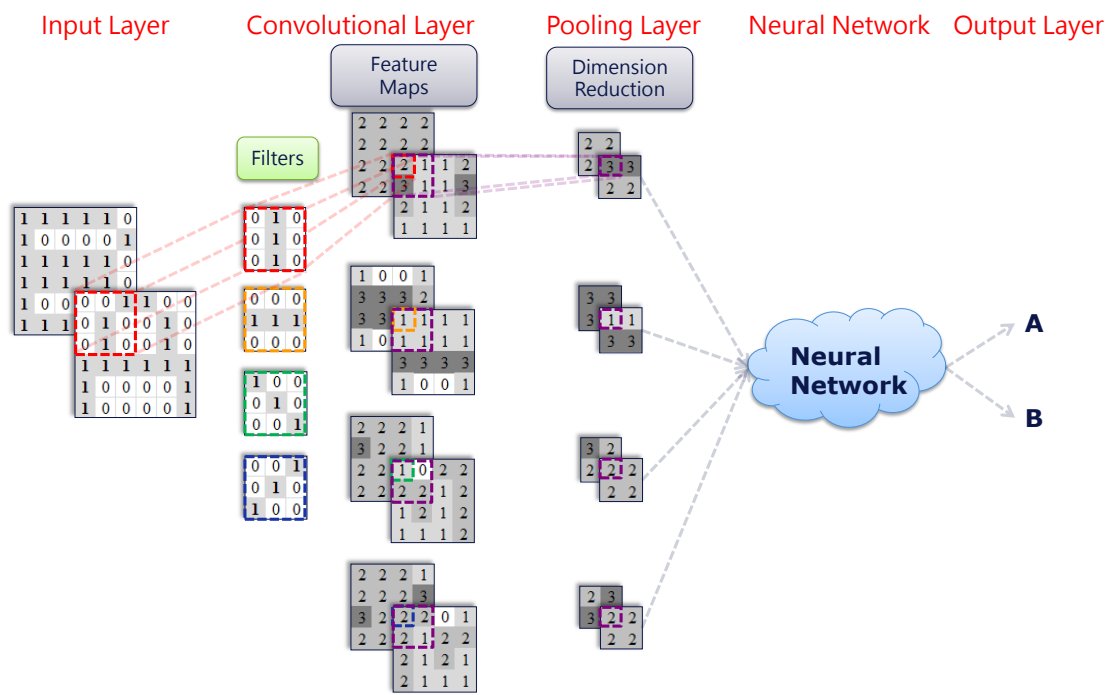

Figure 8. The model of convolutional neural network

**(2). Pooling layers**

In pooling layers, the maximum value and mean value can be calculated for dimension reduction.

**(3). Neural network**

After the computation of convolutional layers and pooling layers, the extracted feature maps can be adopted as the input layer of a neural network, and the future traffic information of target road segment can be adopted as the output layer of the neural network for learning the relationships of feature maps and the future traffic information.

## 3. Travel Time Prediction System and Method

This study proposes a travel time prediction system and describes each component of the system in Subsection 3.1. A travel time prediction method based on convolutional neural networks is presented in Subsection 3.2.

### 3.1. Travel Time Prediction System

The proposed travel time prediction system includes (1) eTag detectors, (2) a traffic control cloud system, (3) a traffic information cloud system, and (4) mobile applications (shown in Figure 9). Each component of system is illustrated as following subsections.

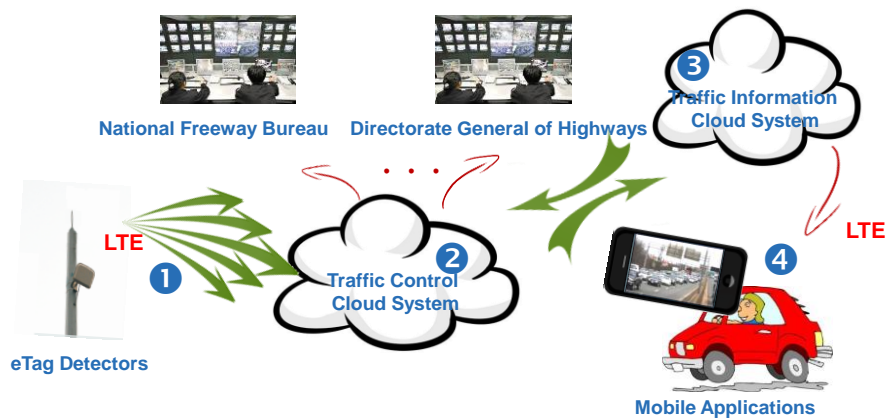

Figure 9. The proposed travel time prediction system

#### 3.1.1. eTag Detectors

The eTag detectors are set up in target road segments to detect the eTag of each vehicle, and the ID of eTag and timestamp are transmitted to the traffic control cloud system via long term evolution (LTE) networks for real-time traffic information estimation.

#### 3.1.2. Traffic Control Cloud System

The traffic control cloud system can received the IDs of eTags and timestamps from eTag detectors to analyse the movement of each vehicle for estimating the travel time of each road segment. The estimated travel time can be transmitted to the traffic information cloud system, National Freeway Bureau, and Directorate General of

Highways for predicting future traffic condition and performing traffic control strategies.

### 3.1.3. Traffic Information Cloud System

The traffic information cloud system can received the real-time travel time from the traffic control cloud system, and the real-time travel time and historical travel time can be adopted into convolutional neural networks to predict the travel time of each road segment. The predicted travel time can be transmitted to mobile applications for user decision-making.

### 3.1.4. Mobile Applications

The mobile applications can be set up in a mobile device and receive real-time and future traffic information from the traffic information cloud system via LTE networks.

## 3.2. Travel Time Prediction Method

This subsection presents the concepts of the proposed model in Subsection 3.2.1. The comparisons of the proposed model and fully-connected neural network are given and discussed in Subsection 3.2.2. Subsection 3.2.3 shows the proposed travel time prediction method.

### 3.2.1. The Concepts of the Proposed Model

In the light of limitations of regression methods and neural networks, this study proposes a time-space travel time model based on a CNN to analyse the interaction effects of space factors and features for obtaining predicted travel time records.

For instance, a $3 \times 3$ dimension model can be designed for the case of Figure 5 (shown in Figure 10). The travel time of the $i$-th road segment at Time $j$ is defined as $T_{i,j}$. In this case, the travel time of the current timestamp and the last 2 timestamps of each road segment in Figure 5 are collected as input parameters and transformed as a $3 \times 3$ dimension model. Furthermore, a $2 \times 2$ dimension filter is designed for the convolutional layer in a CNN (shown in Figure 11). Therefore, the neural network structure of the CNN is constructed as Figure 12. The output (i.e., the estimated travel time of the $i$-th road segment at Time $j+1$) can be estimated as $T_{i,j+1}$' by Equation (12) if the linear model is adopted as the activation function. The parameters $b_{1,1}$, $b_{1,2}$, $b_{1,3}$, $b_{1,4}$, and $b_{2,1}$ are the biases in the CNN. In the application of travel time prediction, the Road Segment 2 may be congestion at Time $j$-1 (i.e., the value of $T_{2,j-1}$ is higher) when the Road Segment 3 may be congestion at Time $j$-2 (i.e., the value of $T_{3,j-2}$ is higher); the Road Segment 1 may be congestion at Time $j$ (i.e., the value of $T_{1,j}$ is higher) when the Road Segment 2 may be congestion at Time $j$-1 (i.e., the value of $T_{2,j-1}$ is higher). Therefore, the interaction effects of space factors and features (i.e., like a slope shape) may be analysed by the filter.

$$T_{i,j+1}' = \left(\sum_{k=1}^{4} w_{2,k} \times h_k\right) + b_{2,1} \text{ where}$$
$$h_1 = w_{1,1} \times T_{1,j-2} + w_{1,2} \times T_{2,j-2} + w_{1,3} \times T_{1,j-1} + w_{1,4} \times T_{2,j-1} + b_{1,1},$$
$$h_2 = w_{1,1} \times T_{2,j-2} + w_{1,2} \times T_{3,j-2} + w_{1,3} \times T_{2,j-1} + w_{1,4} \times T_{3,j-1} + b_{1,2},$$
$$h_3 = w_{1,1} \times T_{1,j-1} + w_{1,2} \times T_{2,j-1} + w_{1,3} \times T_{1,j} + w_{1,4} \times T_{2,j} + b_{1,3},$$
$$h_4 = w_{1,1} \times T_{2,j-1} + w_{1,3} \times T_{2,j-1} + w_{1,3} \times T_{2,j} + w_{1,4} \times T_{3,j} + b_{1,4}$$
(12)

This study used the gradient descent method for the optimization of each weight in the CNN. The loss function ($F$) is defined as a mean square error function (shown in Equation (13)). In each iteration, each weight can be updated in accordance with the learning rate ($\eta$) and the partial differential equation. For instance, the parameter $w_{2,k}$ can be updated by Equation (14) and the parameter $b_{2,1}$ can be updated by Equation (15) for the output layer of the CNN. Furthermore, the parameters of the filter in the convolutional layer of the CNN can be updated by Equations (16), (17), (18), and (19). Each bias in the convolutional layer of the CNN can be updated by Equations (20), (21), (22), and (23).

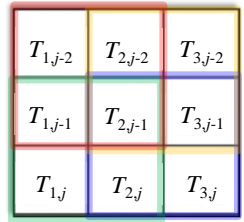

Figure 10. The proposed time-space travel time model: a case study of a 3x3 dimension model

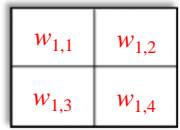

Figure 11. The filter for the convolutional layer in a CNN: a case study of a 2x2 dimension filter

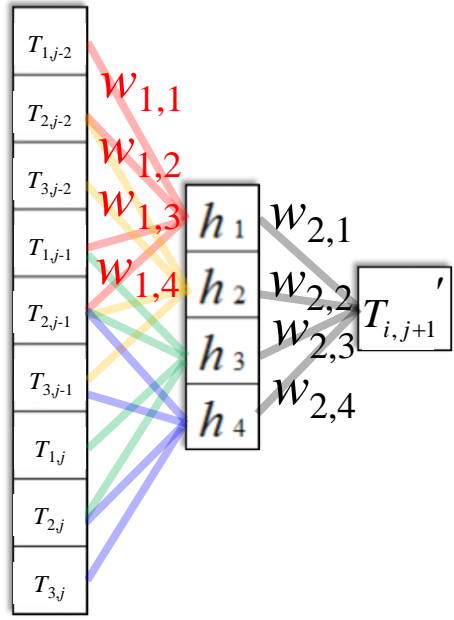

Figure 12. The network structure of CNN

$$F = \frac{1}{2}\left(T_{i,j+1}' - T_{i,j+1}\right)^2 \tag{13}$$

$$w_{2,k} = w_{2,k} - \eta \frac{\partial F}{\partial w_{2,k}} = w_{2,k} - \eta \times [\delta \times h_i] \tag{14}$$

$$b_{2,1} = b_{2,1} - \eta \frac{\partial F}{\partial b_{2,1}} = b_{2,1} - \eta \times \delta \tag{15}$$

$$\begin{aligned} w_{1,1} &= w_{1,1} - \eta \frac{\partial F}{\partial w_{1,1}} \\ &= w_{1,1} - \eta \times [\delta \times (w_{2,1} \times T_{1,j-2} + w_{2,2} \times T_{2,j-2} + w_{2,3} \times T_{1,j-1} + w_{2,4} \times T_{2,j-1})] \end{aligned} \tag{16}$$

$$\begin{aligned} w_{1,2} &= w_{1,2} - \eta \frac{\partial F}{\partial w_{1,2}} \\ &= w_{1,2} - \eta \times [\delta \times (w_{2,1} \times T_{2,j-2} + w_{2,2} \times T_{3,j-2} + w_{2,3} \times T_{2,j-1} + w_{2,4} \times T_{3,j-1})] \end{aligned} \tag{17}$$

$$\begin{aligned} w_{1,3} &= w_{1,3} - \eta \frac{\partial F}{\partial w_{1,3}} \\ &= w_{1,3} - \eta \times [\delta \times (w_{2,1} \times T_{1,j-1} + w_{2,2} \times T_{2,j-1} + w_{2,3} \times T_{1,j} + w_{2,4} \times T_{2,j})] \end{aligned} \tag{18}$$

$$\begin{aligned} w_{1,4} &= w_{1,4} - \eta \frac{\partial F}{\partial w_{1,4}} \\ &= w_{1,4} - \eta \times [\delta \times (w_{2,1} \times T_{2,j-1} + w_{2,2} \times T_{3,j-1} + w_{2,3} \times T_{2,j} + w_{2,4} \times T_{3,j})] \end{aligned} \tag{19}$$

$$b_{1,1} = b_{1,1} - \eta \frac{\partial F}{\partial b_{1,1}} = b_{1,1} - \eta \times [\delta \times w_{2,1}] \qquad (20)$$

$$b_{1,2} = b_{1,2} - \eta \frac{\partial F}{\partial b_{1,2}} = b_{1,2} - \eta \times [\delta \times w_{2,2}] \qquad (21)$$

$$b_{1,3} = b_{1,3} - \eta \frac{\partial F}{\partial b_{1,3}} = b_{1,3} - \eta \times [\delta \times w_{2,3}] \qquad (22)$$

$$b_{1,4} = b_{1,4} - \eta \frac{\partial F}{\partial b_{1,4}} = b_{1,4} - \eta \times [\delta \times w_{2,4}] \qquad (23)$$

### 3.2.2. The Comparisons of the Proposed Model and Fully-Connected Neural Network

In the application of travel time prediction, the traffic information may be influenced by the adjacent road segments and the adjacent time slots. Therefore, the proposed time-space travel time model can be applied to extract the interaction effects of space factors and features by CNN.

In a fully-connected NN (shown in Figure 13), this NN assumes that some interaction effects among all parameters. However, the traffic information may be not influenced by the far road segments and the far time slots. For instance, the travel time $T_{1,j}$ maybe not influenced by the travel time $T_{3,j-2}$ in the case mentioned in Subsection III.B.1. Therefore, some disturbances may occur in a fully-connected NN. For this reasoning, the proposed time-space travel time model is suitably applied as the input layer of CNN for travel time prediction.

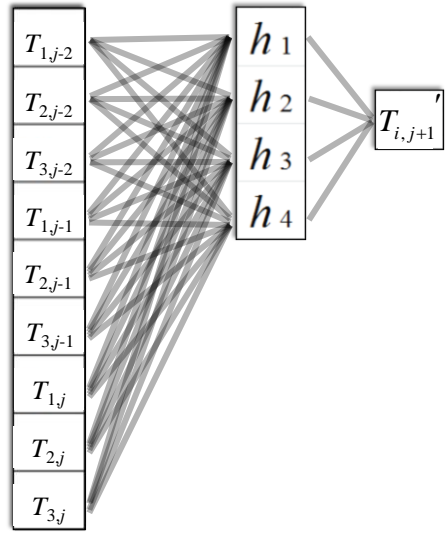

Figure 13. The network structure of a fully-connected NN

### 3.2.3. The Proposed Travel Time Prediction Method

This study adopts a $(x) \times (y+1)$ dimension model into the proposed model for travel time prediction (shown in Figure 14). In this method, *x* road segments, current timestamp, and the last *y* timestamps are considered, so the dimension of input parameters include $(x) \times (y+1)$ travel time records which are adopted into convolutional neural networks for feature extraction. In this case, the travel time of the *i*-th road segment at Time *j* is defined as $T_{i,j}$, and the predicted travel time of the target road segment (i.e., *z*-th road segment) at Time *j*+1 is defined as $T_{z,j+1}$'. The convolutional layers in CNNs can extract the features of travel time relationships between adjacent road segments and adjacent time intervals by filters. The weights in each filter can be generated and optimized by a gradient descent method. Finally, the extracted features are adopted into neural network to learn the relationship between the extracted features and output parameters (i.e., predicted travel time records) for obtaining future travel time.

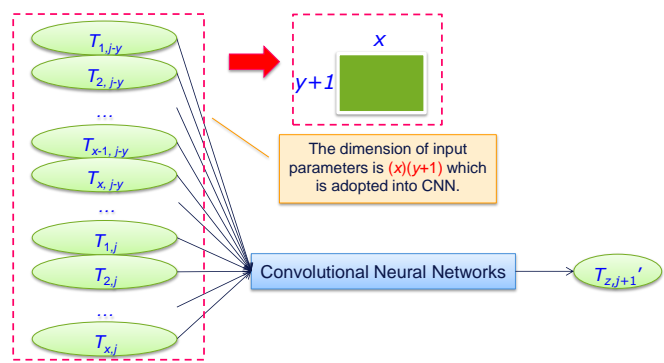

Figure 14. The proposed travel time prediction method

## 4. Practical Experimental Results and Discussions

The practical experimental environments are presented in Subsection 4.1, and the evaluation and comparison of travel time prediction methods are illustrated in Subsection 4.2.

### 4.1. Experimental Environments

In experimental environments, the movement records of each vehicle driven in No. 5 Highway and the alternative roads (i.e., No. 2 Provincial Road, No. 2-C Provincial Road, and No. 9 Provincial Road) were collected by eTag detectors (shown in Figure 15). The travel time records according to the movement records can be obtained in each five-minute interval. Three road segments (i.e., (1) the road segment between Nangang and Toucheng in No. 5 Highway, (2) the road segment between Nuannuan and Toucheng in No. 2 Provincial Road and No. 2-C Provincial Road, and (3) the road segment between Rueibin and Toucheng in No. 2 Provincial Road) in the road map were collected and analysed. For the implementation of the proposed method, the tools of TensorFlow and Keras were used to generate the network structure of NN and CNN. There are two convolutional layers with a $2\times 2$ dimension filter and three hidden fully-connected layers in the CNN. The sigmoid function is applied as the activation function in NN and CNN.

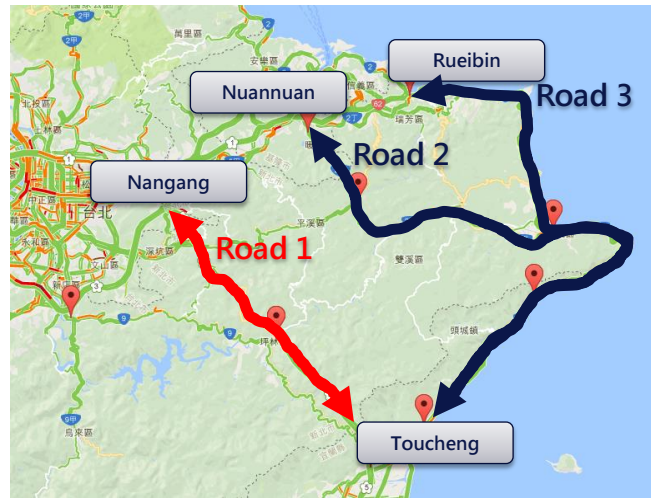

Figure 15. Roads in the experiments

The data from July to December in 2017 were collected and used to evaluate the proposed method. The data from July and November in 2017 were used as training data to train the weights and biases in neural networks, and the data during December in 2017 were used as testing data to test the mean absolute percentage error (MAPE)(shown in Equation (24)) of predicted travel time by the proposed method.

$$e = \frac{\left|T_{z,j+1} - T_{z,j+1}'\right|}{T_{z,j+1}} \tag{24}$$

### 4.2. Analyses and Discussions

The evaluation results of the proposed method for each road segment were showed in Tables 2 and 3. Table 2 showed that the average MAPEs of the proposed method for each road segment in No. 5 Highway was about 5.37%. The average MAPEs was lower than the requirement (i.e., 10%) defined by Ministry of Transportation and Communications (MOTC) in Taiwan [21]. Furthermore, the average MAPEs of the proposed method for each road segment in the alternative roads of No. 5 Highway (i.e., No. 2 Provincial Road, No. 2-C Provincial Road, and No. 9 Provincial Road) was about 5.85% which was lower than the requirement (i.e., 25%) defined by MOTC in Taiwan [21]. Moreover, linear regression, neural network, and convolutional neural networks were implemented and compared in Table 3. The practical results showed that the MAPEs of linear regression, neural network, and the proposed method were 10.04%, 7.42%, and 5.69%, respectively. These results illustrated that the dependency among input parameters were not considered by linear regression, so the MAPE of travel time prediction by linear regression was large. Moreover, some space relationships were between these data, so this study applied convolutional neural networks with several filters to extract important features and improve the accuracy of travel time prediction.

Table 2. The mean absolute percentage error of the proposed method for each road segment

| Road | From | To | The MAPE of travel time prediction |
|---|---|---|---|
| The south direction of Road 1 | Nangang | Toucheng | 7.04% |
| The north direction of Road 1 | Toucheng | Nangang | 3.70% |
| The south direction of Road 2 | Nuannuan | Toucheng | 3.57% |
| The north direction of Road 2 | Toucheng | Nuannuan | 7.34% |
| The south direction of Road 3 | Rueibin | Toucheng | 3.97% |
| The north direction of Road 3 | Toucheng | Rueibin | 8.53% |

Table 3. The mean absolute percentage error of each method

| Method | The MAPE of travel time prediction |
|---|---|
| The requirement defined by Ministry of Transportation and Communications in Taiwan for provincial roads | 25.00% |
| Linear regression | 10.04% |
| Neural network | 7.42% |
| The proposed method based on CNNs | 5.69% |

## 5. Conclusions and Future Work

This section concludes the contributions of this study and discusses the future work.

### 5.1. Conclusions

This study proposes a travel time prediction method based on convolutional neural networks to extract important factors for the improvement of traffic information prediction. In experimental environments, the travel time records of No. 5 Highway and the alternative roads of its from July to December in 2017 were collected and used to evaluate the proposed method. The practical results showed that the MAPE of the proposed method was about 5.69% which is lower than the MAPEs of other methods (e.g., linear regression and neural network). Therefore, the proposed method based on deep learning techniques can improve the accuracy of travel time prediction.

### 5.2. Future Work

Although the results showed that the MAPE of the proposed method was lower than the MAPEs of other methods, the errors of predicted travel time for some road segments were large. Therefore, the unsupervised learning techniques and recurrent neural networks can be considered and applied to perform dimension reduction and time series analyses for the improvement of travel time prediction.

### Data Availability

The data used to support the findings of this study are available from the corresponding author upon request.

### Conflicts of Interest

The authors declare that they have no conflicts of interest.

### Acknowledgment

This work was partially supported by the National Natural Science Foundation of China (Nos. 61906043, 61877010, 11501114, and 11901100), Fujian Natural Science Funds (No. 2019J01243), Funds of Education Department of Fujian Province (No. JAT190026) and Fuzhou University (Nos. 0330/50009113, 510872/GXRC-20016, 510930/XRC-20060, 510730/XRC-18075, 510809/GXRC-19037, 510649/XRC-18049, and 510650/XRC-18050).